\documentclass[conference, 10pt]{IEEEtran}
\IEEEoverridecommandlockouts

\usepackage{times}
\usepackage{epsfig}
\usepackage{graphicx}
\usepackage{amsmath}
\usepackage{amssymb}
\usepackage{cite}
\usepackage{soul}

\usepackage[breaklinks=true,bookmarks=false]{hyperref}

\usepackage{siunitx}

\newcommand\fref{Fig.~\ref}
\newcommand\tref{Table~\ref}
\newcommand\sref{Sec.~\ref}

\def\0{{\bf 0}}

\usepackage{amsthm}

\usepackage{amsthm}

\newtheorem{mytheorem}{Theorem}
\newtheorem{assumption}[mytheorem]{Assumption}
\newtheorem{lemma}[mytheorem]{Lemma}
\newtheorem{remk}[mytheorem]{Remark}

\usepackage{amsmath,amsfonts,bm}

\usepackage{xcolor}
\usepackage{colortbl}

\def\eqref#1{Eqn.~\ref{#1}}

\def\1{\bm{1}}

\DeclareMathAlphabet{\mathsfit}{\encodingdefault}{\sfdefault}{m}{sl}
\SetMathAlphabet{\mathsfit}{bold}{\encodingdefault}{\sfdefault}{bx}{n}

\usepackage{multirow}
\usepackage{color}
\usepackage{xcolor}
\usepackage{amsmath}
\definecolor{Gray}{gray}{0.85}

\usepackage[ruled,noend]{algorithm2e}

\SetCommentSty{mycommfont}

\SetKwInput{KwInput}{Input}

\usepackage{booktabs} %
\usepackage{multicol}
\usepackage{diagbox}

\newcommand{\OURS}[0]{HAO\xspace}

\def\BibTeX{{\rm B\kern-.05em{\sc i\kern-.025em b}\kern-.08em
    T\kern-.1667em\lower.7ex\hbox{E}\kern-.125emX}}
\begin{document}

\setlength{\parskip}{-0.00cm}
\setlength{\dblfloatsep}{0.2cm}
\setlength{\dbltextfloatsep}{0.2cm}
\setlength{\floatsep}{0.2cm}
\setlength{\textfloatsep}{0.183cm}

\title{\OURS: Hardware-aware Neural Architecture Optimization for Efficient Inference}
\author{Zhen Dong$^*$\thanks{$^*$ Equal contribution.}, Yizhao Gao$^*$, Qijing Huang, John Wawrzynek, Hayden K.H. So, Kurt Keutzer\\
University of California, Berkeley\\
The University of Hong Kong\\
{\tt\small \{zhendong,qijing.huang,johnw,keutzer\}@berkeley.edu, \{yzgao,hso\}@eee.hku.hk}

}
\maketitle

\thispagestyle{plain}
\pagestyle{plain}

\begin{abstract}
Automatic algorithm-hardware co-design for DNN has shown great success in improving the performance of DNNs on FPGAs. 
However, this process remains challenging due to the intractable search space of neural network architectures and hardware accelerator implementation.
Differing from existing hardware-aware neural architecture search (NAS) algorithms that rely solely on the expensive learning-based approaches, our work incorporates integer programming into the search algorithm to prune the design space.
Given a set of hardware resource constraints, our integer programming formulation directly outputs the optimal accelerator configuration for mapping a DNN subgraph that minimizes latency.
We use an accuracy predictor for different DNN subgraphs with different quantization schemes and generate accuracy-latency pareto frontiers.
With low computational cost, our algorithm can generate quantized networks that achieve state-of-the-art accuracy and hardware performance on Xilinx Zynq (ZU3EG) FPGA for image classification on ImageNet dataset. 
The solution searched by our algorithm achieves 72.5\% top-1 accuracy on ImageNet at framerate 50, which is 60\% faster than MnasNet~\cite{tan2019mnasnet} and 135\% faster than FBNet~\cite{wu2019fbnet} with comparable accuracy.
\end{abstract}

\section{Introduction}
\label{sec:intro}
Modern complex deep neural networks (DNNs) are able to achieve unparalleled accuracy in a wide range of applications at the expense of their much increased computing requirements.
To successfully deploy these computationally demanding DNNs in resource-constrained edge systems such as an embedded FPGA, while maintaining real-time performance, system designers must therefore engage in difficult tradeoffs between model accuracy and implementation efficiency.
There are three common approaches to improve the efficiency of DNN and the corresponding hardware design for edge deployment:
1) quantize the model to achieve efficient representations of DNNs, 
2) select less compute-intensive operations and design efficient DNN architectures, 
and 3) design specialized hardware.
The three design techniques altogether form a large design space for developing efficient DNN accelerator solutions at the edge.

Quantization~\cite{courbariaux2016binarized, jacob2018quantization, zhang2018lq, zhou2017incremental} is a general and effective technique that uses low bitwidth (such as 4-bit or 8-bit) to represent the floating-point weights/activations in neural networks. To achieve a better trade-off between accuracy and efficiency, mixed-precision quantization was introduced to allow different layers to have different bitwidths. 
Mixed-precision quantization leads to an exponentially large search space to find the optimal bitwidths. Prior work~\cite{wu2018mixed, wang2019haq, dong2019hawq} adopts differentiable search, reinforcement learning, or sensitivity analysis to tackle this problem. However, the computational cost of these approaches is non-trivial. Besides, these works solely focus on quantization without co-considering the neural architecture design or efficient hardware implementation. 

The second approach to achieve efficient inference is designing compact neural network architecture. Compared to manually designing networks, neural architecture search (NAS) algorithms~\cite{zoph2016neural, liu2018progressive, liu2018darts, tan2019mnasnet, cai2018proxylessnas} can automatically find network architectures that are more accurate and efficient. However, the NAS algorithm typically requires training sampled networks/sub-networks to get feedback on different neural architectures, 
which makes NAS algorithms computationally expensive to gain enough feedback and achieve good performance. In practice, NAS algorithms either heuristically prune the architectural search space or use proxy tasks to reduce the computational cost, leading to sub-optimal DNN architectures.

The hardware design, in common practice, is performed separately with software.
Such practice can lead to sub-optimal performance because quantization and NAS target hardware-agnostic metrics such as model size or FLOPs. These performance proxies do not guarantee high inference speed on different hardware designs.
As an example, the quantization algorithm may select a mixture of every bitwidth from 1 bit to 8 bit, and the NAS algorithm may choose to jointly use convolution with different kernel and group sizes.
Though this solution can be small in model size or FLOP counts, on embedded FPGA devices with limited resources (such as Zynq ZU3EG), supporting all these operations at the same time is inefficient or even infeasible.

Consequently, a joint search among quantization, neural architectures, and hardware implementation is necessary to expose the optimal configurations of DNN and the corresponding accelerator design.  
Previous work \cite{wang2019haq, wu2018mixed, yao2020hawqv3} searched quantization schemes with different hardware configurations, but left their DNN architecture untouched. 
\cite{tan2019mnasnet, wu2019fbnet, cai2018proxylessnas} searched for efficient neural architectures on specific hardware platforms, but did not consider the impact of quantization and hardware design.
\cite{jiang2020hardware, hao2019fpga, hao2019nais, abdelfattah2020best} covered hardware design and neural architectures in their search space but did not include quantization. Though \cite{jiang2020standing, lu2019neural, li2020edd, huang2021codenet} considered all three aspects, their search space is limited.

In this work, we explore the joint search space of neural architecture, quantization, and hardware design. 
Instead of pruning the space by heuristics, or applying reinforcement learning or derivative-based search algorithms, we formulate the search as an integer programming problem, so that efficient optimization algorithms can be used to reduce computational cost. 
Based on our hardware latency model and network accuracy predictor, we propose a \textbf{h}ardware-aware neural \textbf{a}rchitecture \textbf{o}ptimization (\OURS) method
to generate pareto-optimal DNN designs to run on embedded FPGAs.
Our contributions are as follows:
\begin{enumerate}
    \item We formulate the design of neural architecture, quantization, and hardware design jointly as an integer programming problem.
    \item We use a subgraph-based latency model for FPGAs, and we use a network accuracy predictor to reduce the computational cost of the automatic design flow. 
    \item \OURS achieves state-of-the-art performance on ImageNet with Zynq ZU3EG FPGA. Our model can achieve 72.5\% Top-1 accuracy running with 50 FPS, which is 60\% faster than MnasNet and 135\% faster than FBNet with comparable accuracy.
\end{enumerate}

\section{Related Work}
\label{sec:related_work}

\subsection{Efficient Deep Learning}
Quantization \cite{courbariaux2016binarized, jacob2018quantization, zhang2018lq, zhou2017incremental, wang2019haq, dong2019hawq, cai2020zeroq} is a practical method to achieve efficient inference, which uses low bitwidth to represent weights and activations in a given neural network model. Since uniformly applying ultra-low precision quantization can cause accuracy degradation, mixed-precision quantization \cite{zhou2018adaptive, wang2019haq, dong2019hawqv2} is used to recover the accuracy. Mixed-precision quantization allows different layers in a neural network to have different bitwidth, leading to an exponentially large search space for the optimal bitwidth setting. \cite{wang2019haq} applies reinforcement learning to explore the space, and \cite{wu2018mixed} uses differentiable search to decrease the required search time. \cite{dong2019hawqv2} introduces Hessian-based sensitivity analysis to determine bitwidth, while obtaining the Hessian information has a high computational cost.

Instead of compressing a large pre-trained model, previous work \cite{iandola2016squeezenet, sandler2018mobilenetv2, ma2018shufflenet, tan2019efficientnet} focus on directly designing compact neural network architectures that can achieve decent accuracy with small model size or FLOPs. To avoid manual efforts, neural architecture search (NAS) algorithms have been proposed to automatically design pareto-optimal network architectures. Previous NAS methods \cite{zoph2016neural, liu2018progressive} use a reinforcement learning agent to explore the design space of neural architectures, which typically requires a large number of computational resources (48,000 GPU hours). \cite{real2019regularized} applies evolutionary algorithm to search for efficient neural architectures, which is feasible but also costly (75,600 GPU hours). Differential search based NAS methods \cite{liu2018darts, wu2019fbnet, tan2019mnasnet, cai2018proxylessnas} significantly reduce the search cost by 1) using a supernet with weight sharing \cite{pham2018efficient} and 2) applying continuous relaxation on the discrete search space so that gradients can be used to assist searching. However, differentiable NAS algorithms often lead to a small search space due to the limitation of supernets, which makes it dependent on existing candidates of good operations. They are sub-optimal if the design space is not already well-explored.

\subsection{Hardware-aware Search}
Since inference speed is dependent on characteristics of specific hardware platforms, simply applying quantization or NAS algorithms based on proxy metrics (model size or FLOPs) can be sub-optimal. To solve this problem, many hardware-aware search algorithms have been introduced to seek efficient deployment of DNNs on targeted hardware platforms. These methods \cite{cai2018proxylessnas, wang2019haq, cai2019device,tan2019mnasnet, wu2019fbnet, scheidegger2019constrained, jiang2019accuracy, yang2019fpnet, blott2018finn} usually retrieve latency or energy feedback from a given hardware platform, and search for optimal DNNs that can meet certain application constraints. 
Note that the hardware design is fixed in these methods, and thus is not in the search space.

To further improve the efficiency, in recent years, a few works have extended the NAS framework by integrating hardware design into the search space \cite{zhang2019neural,lu2019neural, hao2019fpga, hao2019nais, li2020edd, jiang2020standing, jiang2020hardware, yang2020co, abdelfattah2020codesign, abdelfattah2020best}. %
Generally, these software/hardware co-search algorithms adopt pre-defined hardware design templates and incorporate several high-level design hyperparameters in the search framework.
In addition to neural architectures, some works also incorporated quantization in their search space. %
\cite{lu2019neural} captures the relationship between quantization bitwidth and LUTs consumption on FPGA, and developed a NAS algorithm under the constraint of LUTs. In \cite{jiang2020standing}, the authors integrate several model compression techniques in the search framework and use quantization to reduce the latency of weight loading. \cite{li2020edd} proposes a uniformed differentiable search algorithm using gumbel-softmax to sample discrete implementation hyperparameters including quantization bitwidth. 

Although previous methods consider hardware design choices, the size of searchable space is still limited by the search algorithm efficiency and the total computation budget. Consequently, enlarging hardware search space may result in the shrinkage of software search space.
In this work, we propose a subgraph-based hardware latency model, together with an accuracy predictor for neural architectures and quantization. Based on these, we are able to formulate the software/hardware co-search as an integer programming problem, which can be effectively optimized with a very small computational cost.

\section{Methodology}
\label{sec:methodology}

In \OURS, we expose a large design space in both hardware and algorithm configurations to accelerate DNNs. 
To efficiently navigate the search space, we first apply integer programming to prune the hardware configuration space by minimizing the latency subject to a set of hardware resource constraints. We then narrow the DNN architecture space by adopting Monte Carlo tree search (MCTS)~\cite{kocsis2006bandit} to minimize the quantization accuracy perturbation while satisfying a given latency constraint. 
In addition, we develop an accuracy predictor to estimate the accuracy of the DNN to further reduce the overall feedback time for each sample. 
Our flow produces a pareto-optimal curve between latency and accuracy. 
\subsection{Hardware Design}
\label{subsec:hardware_design}
We target FPGA in this work to demonstrate how co-designed hardware and DNN fully exploit the optimization opportunities in hardware with limited resources while achieving on-par accuracy.
In this section, we model the resource consumption and the computation latency for different types of convolution kernels. On top of that, we formulate the overall resource constraints and latency objectives as an integer programming problem for the subgraph-based design, which will serve as the latency simulator in the following DNN architecture optimization.

\subsubsection{Hardware Subgraph Template}
    \begin{figure}[]
    \centering
    \includegraphics[width=.45\textwidth]{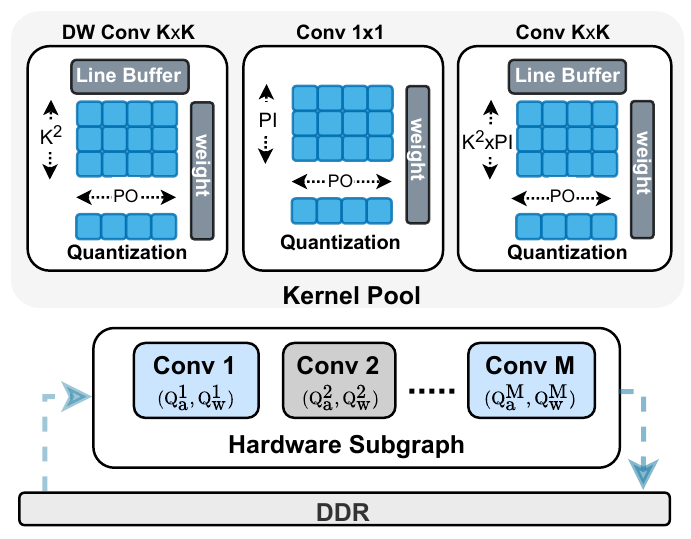}
    \vspace{-1mm}
    \caption{
    \small
      Hardware design space. The dataflow accelerator template consists of $M$ convolution kernels that are selected from the kernel pool and spatially mapped to hardware. The tunable design parameters include the number of compute kernels $M$,  the kernel type, filter size $K$, input and output channel parallelization factor $PI$ and $PO$.  
}
      \label{fig:hardware_subgraph}
    \end{figure}
As shown in ~\fref{fig:hardware_subgraph}, in \OURS, we adopt a subgraph-based hardware design. A subgraph consists of several convolution kernels that are spatially mapped on hardware, which also corresponds to the major building block of neural architecture.
For a given hardware subgraph, the possible building blocks for neural architecture also include all the sub-layers of the subgraph since each kernel is implemented with a skip signal to bypass its compute in hardware. %
Each invocation to the accelerator computes one subgraph in the DNN architecture. The intra-subgraph results are buffered and streamed on FPGA and the inter-subgraph activations are communicated through DRAM.

We implement a parameterizable accelerator template in high-level synthesis (HLS). 
The generated dataflow accelerator can contain M convolution kernels chained through FIFOs to exploit pipeline-level parallelism. Each convolution kernel can be chosen from one of the three convolutions from the kernel pool: Conv $k\times k$, Depthwise Conv $k\times k$ \cite{chollet2017xception}, and Conv $1\times 1$. 
The hardware implementation of each kernel typically comprises a weight buffer, a line buffer, a MAC engine, and a quantization unit to rescale outputs.%
All the computational units are implemented using integer-only arithmetics as in \cite{jacob2018quantization}.

\subsubsection{Hardware Resource Modeling}
    \label{sec:hw_resource}
This section describes the modeling details of different FPGA resources. We adopt a bottom-up design flow to model the utilization of LUTs and DSPs for low-bit multiply-accumulate (MAC) operations on FPGA. In addition, our model derives the BRAM utilization based on data size and precisions as well as the parallelization factors of the compute kernels.
Table~\ref{tab:notation} lists the notations used in this paper.

\begin{figure}[t]
    \centering
    \includegraphics[width=.42\textwidth]{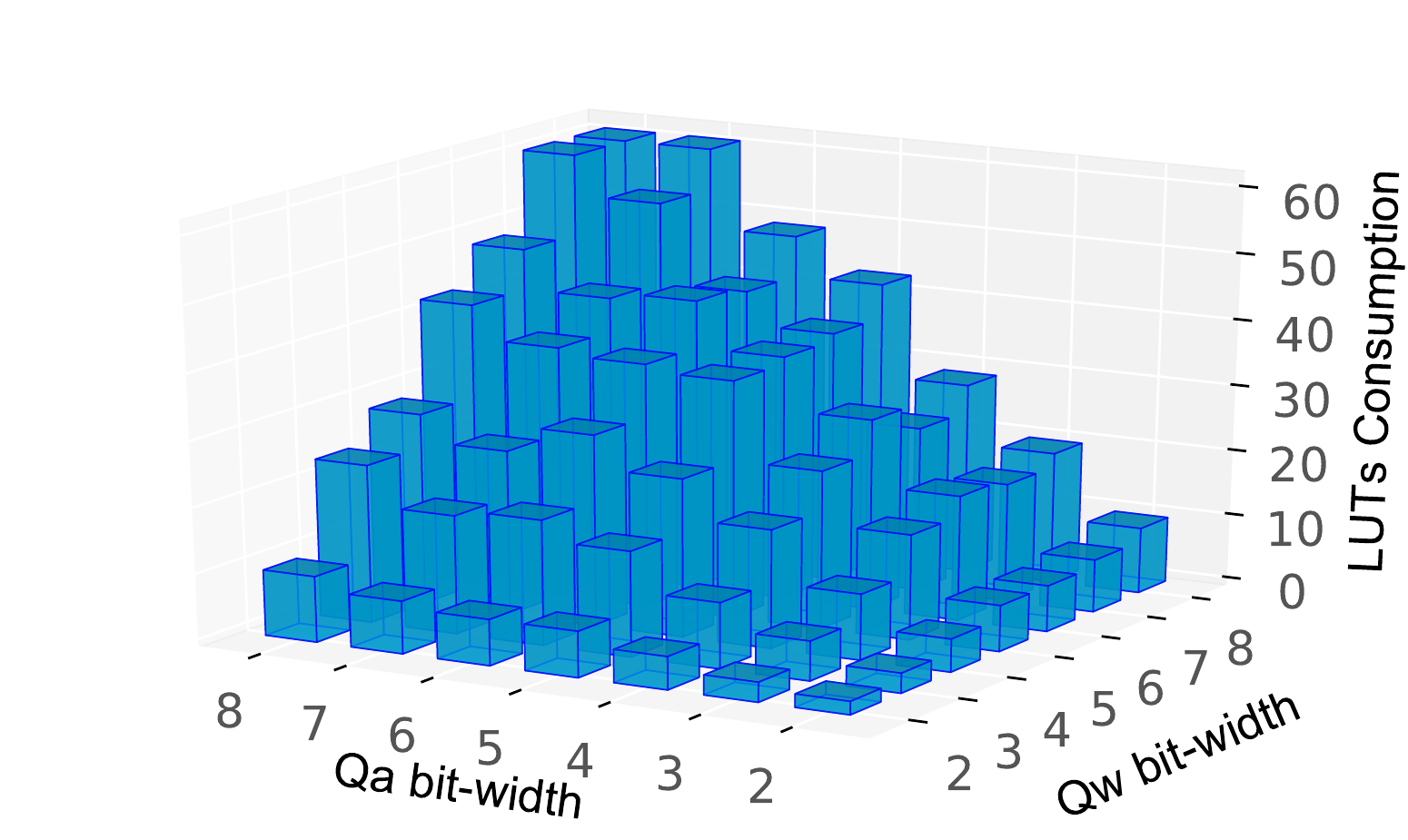}
    \vspace{-1mm}
    \caption{
    \small
      LUT usage of multipliers with different input precisions. }
      \label{fig:luts_multiplier}
    \end{figure}

    \paragraph*{LUTs} Both DSPs and LUTs can be used for  computation on FPGA. It is more efficient to perform ultra low-bit computation on LUTs compared with DSPs. 
    We use pragma to direct the mapping of low-precision MAC operations to LUTs in HLS.
    To build a precise model, we perform full logic synthesis to obtain the LUTs consumption on low bitwidth multipliers and adders.~\fref{fig:luts_multiplier} shows the LUTs consumption on different activation and weight bitwidths ranging from 2 to 8. We denote the LUT resource lookup function of multipliers as $L_M(Q_w, Q_a)$ where $Q_w$ and $Q_a$ represent the bitwidth of weights and input activations respectively.  
    Derived from the logic synthesis results, the LUT consumption of the adders $L_A(Q_p)$ for carrying out $Q_p$ bit partial sum accumulation can be expressed as $L_A(Q_p) = Q_p + 7$.

    \begin{figure}[t]
    \centering
    \includegraphics[width=.40\textwidth]{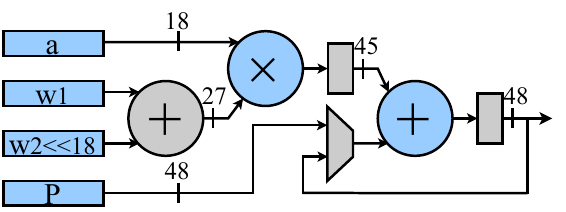}
    \vspace{-1mm}
    \caption{
    \small
    Example mapping of two low-precision MACs $a\times w_1$ and $a\times w_2$ onto a DSP with $27\times18$ multiplier support. 
      The multiplexer in DSP can choose between self-accumulating or chaining mode.
      }
      \label{fig:dsp_diagram}
    \end{figure}

   \paragraph*{DSP} %

   The embedded DSP slice on FPGA supports the MAC operation in the following format:  
    \begin{equation}
    \small
    \label{eq: dsp_func}
        P \mathrel{+}= A \times (B + C)
    \end{equation}
    In naive HLS mapping, one DSP slice is configured to support one MAC. 
    To improve DSP throughput for low-bit operations, we use the shift-and-pack method in \cite{fu2016deep} to efficiently map two MACs on one DSP by leveraging the additional pre-adder. Given the input activation $a$ and the weights $w_1$ and $w_2$ for two different output channels, as shown in Fig.~\ref{fig:dsp_diagram}, 
    the packing algorithm first sign-extends $w_1$ to 27 bits and left shifts $w_2$ by $18$ bits. The output $P$ can be further accumulated with the partial sum or separated into two products $P_1$ and $P_2$. This shift-and-pack method can be applied to the situation when $w_1$ and $w_2$ are no larger than 8 bits. %

    \paragraph*{BRAM}
    We assume a buffering scheme in which we fully exploit reuse opportunities.
    The 18-Kb BRAMs usage $B_w$ for the weight buffer can be calculated as: 
    
    \begin{equation}
        \label{eq:bram_weigt}
        \small
         B_w = \lceil N_w \times Q_w / PF / 18 {\rm{Kb}} \rceil \times PF 
    \end{equation}
    where $N_w$ is the maximum number of weights to store on-chip, $Q_w$ is the bitwidth of weights, and $PF$ is the BRAM partition factor of the weights buffer. 
    For convolution kernel with size $k > 1$, we implement a line-buffer to maximize input reuse.
    The number of BRAMs $B_l$ needed for line buffer is:
    \begin{equation}
        \label{eq:bram_lineb}
        \small
        B_l = \lceil (W \times C)_{\max} \times Q_a /18{\rm{Kb}}\rceil \times k       
    \end{equation}
    where $(W \times C)_{max}$ is the maximum product between the size of image width $W$ and channel $C$ over the entire network. Our line buffer implementation merges the input width and channel dimension of the feature map into one dimension, and $k$ rows of line buffers are allocated for $k \times k$ convolution kernel.

    \subsubsection{Hardware Resource Allocation} 
    With the resources modeling, we can further estimate the optimal resource allocation for a hardware subgraph under the resource constraints of the target FPGA.
    For full \textit{k $\times$ k Conv}, given the input channel parallelization factor $PI$ and output channel parallelization factor $PO$, the compute engine loads $k^2 \times PI$ inputs in parallel and computes $PO$ output partial sums. The total BRAM usage $N_{wbuf}$ for on-chip buffers is: 
     
    \begin{equation}
        \small
        \begin{gathered} 
        \begin{aligned}
            N_{\text{wbuf}} &= \left\{
            \begin{aligned}
                B_w &+ B_l   &k > 1 \\ 
                &B_w  &k = 1 \\    
            \end{aligned} 
            \right.\\
        \end{aligned}\\
        \end{gathered}
    \end{equation} 
    
The engine is composed of $ k^2 \times PI \times PO$ MAC units that can be mapped to either DSPs or LUTs, incurring usage in LUTs $N_{luts}$ or DSPs $N_{dsp}$:
    \begin{equation}
        \small
        \begin{gathered} 
        \begin{aligned}
            N_{\text{dsp}} &= k^2 \times PI \times PO /2  \\
        \end{aligned}\\
            N_{\text{luts}} = k^2 \times PI \times PO \times (L_M(Q_w, Q_a) + L_A(Q_p)) \\
        \end{gathered}
    \end{equation}
    
    For \textit{k $\times$ k Depthwise Conv} where each output channel result is corresponding to the inputs from the same channel, we use only $PO$ to denote the channel dimension parallel factor.
    The $k \times k$ computation engine takes $k^2 \times PO$ input and computes $PO$ partial sums concurrently. Similarly, the BRAM usage for the compute kernel is: 
    \begin{equation}
        \small
        \begin{gathered}
            N_{\text{wbuf}} = B_w + B_l \\
        \end{gathered}
    \end{equation}
    The LUT or DSP usage to support depthwise convolution grows linearly with the $PO$ parallelism factor:
    \begin{equation}
        \small
        \begin{gathered}
            N_{\text{dsp}} = k^2 \times PO \\
            N_{\text{luts}} = k^2 \times PO \times (L_M(Q_w, Q_a) + L_A(Q_p)) \\
        \end{gathered}
    \end{equation}    
    Regarding the \textit{Quantization} unit that converts partial sum in high-precision to quantized input for the next layer, we implement it with DSP with a parallelization factor of $PO$. Its overall resource usage is: 
    
    \begin{equation}
        \small
        N_{\text{dsp}} = PO,
        N_{\text{sbuf}} = B_s \\
    \end{equation}
    Since we perform channel-wise quantization on weights, each output channel has its own quantization scale.
    We thus set the number of buffered scales $N_s$ to $OC$.
    The calculation of $B_s$ is similar to $B_w$ in ~\eqref{eq:bram_weigt}. 
    The bitwidth of scale $Q_s$ ranges from 16-24 depending on the actual value range after obtaining the integer scale using the inference scheme in \cite{jacob2018quantization}. 
    The total BRAM usage $N_{\text{bram}}$ is a sum of weight buffer usage $N_{\text{wbuf}}$ and scale buffer usage $N_{\text{sbuf}}$: 

    \begin{equation}
        \small
        N_{\text{bram}} = N_{\text{wbuf}} + N_{\text{sbuf}} \\
    \end{equation}    

    \subsubsection{Hardware Latency Objective}
    
     \begin{table}[t]
    \centering
    \scriptsize
        \caption{Notations for hardware design}
        \vspace{-2mm}
        \begin{tabular}{p{0.6cm}<{\centering}c|p{0.6cm}<{\centering}c}
        \hline
        Notation & Description & Notation & Description\\
        \hline
        $H$   & feature map height     & $PI$      & parallelism on input channel\\
        $W$   & feature map width      & $PO$      & parallelism on output channel \\
        $Q$   & quantization setting   & $PF$      & array partition factor \\
        $Q_a$ & activation bitwidth   & $L_M$     & LUTs usage of a Multiplier\\
        $Q_w$ & weights bitwidth      & $L_A$     & LUTs usage of an Adder \\
        $Q_p$ & partial sum bitwidth  & $B_l$     & line buffer BRAM usage\\
        $k$   & kernel size            & $B_w$     & weights BRAM usage\\
        $Lat_{\text{comp}}$& computation latency & $N_w$  & number of weights buffered\\
        $Lat_{\text{on/off}}$ &latency of activation  & $N_{\text{dsp}}$ & total DSP usage of a kernel \\
        &  communication & $N_{\text{bram}}$ & total BRAM usage of a kernel \\
        $Lat_{w}$ &latency of loading weights & $N_{\text{luts}}$& total LUTs usage of a kernel \\
        $S$   & hardware subgraph & $N_{\text{wbuf}}$  & BRAM usage for weights buffer\\
        $A$  & neural architecture & $N_{\text{sbuf}}$ & BRAM usage for scale buffer\\
        $M$   & number of kernels in $S$ & $N$  & number of layers in $A$ \\
        \hline
        \end{tabular}
        
        \label{tab:notation}
    \end{table}

    Given a layer with input channel size $IC$, output channel size $OC$, input height $H$ and width $W$, the compute latency is:
    \begin{equation}
    \centering
    \small
        Lat_{\text{comp}} = \left\{ 
        \begin{aligned}
        &H \times W \times \lceil IC / PI\rceil  \times \lceil OC / PO \rceil  &~\text{if \textit{full}}\\
        &H \times W \times \lceil IC / PO\rceil &~\text{if \textit{depthwise}} \\
        \end{aligned}
        \right.
    \end{equation}
    depending on if the kernel type is full or depthwise convolution.
    The communication latency for loading the activation on-chip and off-chip can be roughly calculated as:
    \begin{equation}
        \small
        \begin{gathered}
            Lat_{\text{on}} = H \times W \times IC \times Q_a / bw \\
            Lat_{\text{off}} = H \times W \times OC \times Q_a / bw \\ 
        \end{gathered}
    \end{equation}
    where $bw$ is the practical bandwidth of off-chip memory. Similarly, the latency of loading weights can be estimated as:
    \begin{equation}
    \centering
    \small
        Lat_{w} = \left\{ 
        \begin{aligned}
        &k^2 \times IC \times OC \times Q_w / bw   &~\text{if \textit{full}}\\
        &k^2 \times IC \times Q_w / bw &~\text{if \textit{depthwise}} \\
        \end{aligned}
        \right.
    \end{equation}

    Based on the latency model for a single layer, we can further derive the latency of computing a subgraph. A hardware subgraph design with $M$ convolution kernels can be represented as $S = \{K_1, K_2, ...K_M\}$ with specific quantization bitwidths $Q = \{(Q_a^1, Q_w^1),...,(Q_a^M, Q_w^M)\}$. For a given network architecture $A = \{a_1,a_2, ...a_N\}$, the subgraph mapping $\{g_1, ..., g_L\}$ 
   can be generated using a grouping function $f_m$:
    \begin{equation}
    \small
            \{g_1, g_2, ...,g_L\} = f_m(\{a_1,a_2,...,a_N\})\\
    \end{equation}
    To model the overlapping of the dataflow architecture, the latency of computing each $g_i$ can be approximated using the maximum latency over all the subgraph layers. Besides, to execute each layer on hardware, the accelerator will preload the weights to the on-chip buffer before the kernel starts, and apply double-buffering to hide the communication overhead of the input activations. The overall latency for computing a subgraph can be written as:
    \begin{equation}
    \small
    \begin{aligned}
            Lat(g_i) =& \max(Lat_{\text{on}}^{i1}, Lat(a_{i1}),..., Lat(a_{iM}), Lat_{\text{off}}^{iM})\\
        & + \sum\nolimits_{j=1}^M Lat_{w}^{ij}
    \end{aligned}
    \end{equation}

    \begin{figure*}[]
    \centering
    \includegraphics[trim=66 210 80 140, clip, width=0.96\textwidth]{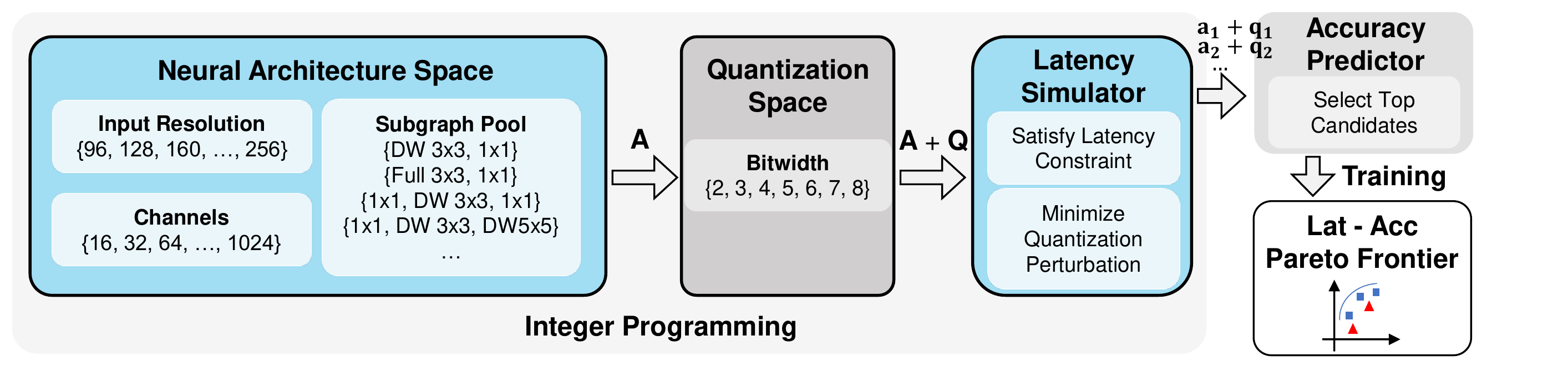}
    \vspace{-3mm}
    \caption{
    \small
    Illustration of \OURS pipeline.}
      \label{fig:software_illustration}
    \end{figure*}
    With the hardware analytical model above, we can then formulate the automatic hardware design problem as an integer programming that minimizes the overall latency:
    \begin{equation}
    \label{eq:hardware_program}
    \small
        \begin{aligned}
        &\text{min} & & \sum\nolimits_{i=1}^L Lat(g_i) \\
        &\text{s.t.} & & \sum_{k \in S} N_{\text{dsp}}^{k} \leq T_{\text{dsp}} \\
        & & &\sum_{k \in S} N_{\text{luts}}^{k} \leq T_{\text{luts}} \times \beta \\
        & & &\sum_{k \in S} N_{\text{bram}}^{k} \leq T_{\text{bram}}
        \end{aligned}
    \end{equation}
    where $T_{\text{dsp}}$, $T_{\text{luts}}$, $T_{\text{bram}}$ are the total resources available on the target FPGA device. Note that $\beta$ is an empirical parameter describing the percentage of total LUTs allocated for MAC computation, which is set to 50\% in our experiments.  
    We treat this formulation as a sub-program to the DNN design optimization which will be covered in the next section.
    Given the explicitly expressed constraints and objective, we are able to directly generate the corresponding hardware implementation that minimizes the latency for different DNN design choices with different quantization schemes and kernel types.

\subsection{DNN Design}
Co-search of hardware-friendly neural network architectures and mixed quantization precisions is computationally intensive and time-consuming. 
In~\OURS, we formulate the search to an integer programming problem.
In~\sref{subsec:search_space} we present our search space of neural architectures. Given a latency constraint, we can first search feasible neural architectures and corresponding mixed-precision bitwidth settings by applying the aforementioned hardware latency model as well as a model quantifying the effect of quantization perturbation. %
We then use an accuracy predictor to compare across different networks and find the pareto-optimal architectures and quantization settings among all candidates. %

\subsubsection{Search Space of Neural Architectures}
\label{subsec:search_space}

In \OURS, we construct the neural network architectures from subgraphs with feasible hardware mappings on FPGAs. 
Our subgraphs are combinations of operations such as convolution or depthwise convolution with kernel size of $1\times1$ or $k\times k$ as mentioned in the previous section.
Although only one subgraph can be chosen on hardware, the possible building blocks for neural architecture search include the sub-layers of the subgraph. This is because each layer in the subgraph can be decided whether to bypass or not using a skip signal in hardware. 

We set no limit on the total number of subgraphs and choose the channel size for different layers from $\{16, 32, 64, 128, 256, 512, 1024\}$. 
We also consider input resolution in \OURS with potential configuration from $\{96, 128, 160, 192, 224, 256\}$.
Consequently, our search space is significantly larger compared to the  prior work \cite{cai2018proxylessnas, liu2018darts, wu2019fbnet, xu2019pc, tan2019mnasnet}. For example, in \cite{tan2019mnasnet}, the same cell configuration is repeated within every block. A standard search setting is to use 5 blocks with 3 identical cells in each block, and each cell, typically with 3 layers,  %
has a sub-search space of 432, resulting in a search space of size $432^{5} \approx 10^{13}$. %
In comparison, even with a simple subgraph \{\textit{1x1 convolution}, \textit{3x3 depthwise convolution}\}, assume the number of layers is $45$ (same as \cite{tan2019mnasnet}), 
the size of search space in \OURS is $(2\times7)^{45} \approx 10^{51}$. 
The large search space of \OURS makes it more likely to encompass designs with good efficiency and high accuracy for broader deployment scenarios with various hardware and latency constraints.  

\subsubsection{Integer Programming}
\label{subsec:integer_programming}
Given a latency constraint $Lat_0$, we use integer programming to obtain feasible neural architectures and corresponding quantization settings. Specifically, based on the aforementioned hardware simulator, inference latency ($Lat$) is a function (denoted as $\mathbb L$) of neural architecture ($A$) and the quantization setting ($Q$) for subgraph. In \eqref{latency_a_q}, i and j are layer index, $N$ represents the total number of layers, and $M$ represents the number of layers in a subgraph.

\begin{equation}
    \small
        \begin{gathered}
        Lat = \mathbb L(A, Q), \\
        A = \{k_i, H_i, W_i, IC_i, OC_i, S_i, i \in [1, N]\}, \\
        Q = \{Q_a^j, Q_w^j, j \in [1, M]\} \\
        \end{gathered}
    \label{latency_a_q}
\end{equation}

In \OURS, perturbation, denoted as $Pert$, is used to estimate the accuracy degradation caused by quantization. For a given neural architecture, the accuracy of the full-precision pretrained model is irrelevant to quantization setting $Q$.
The perturbation models the relative accuracy change to the pertrained network among different $Q$. 
As shown in~\eqref{perturb_a_q}, the perturbation should be multiplied with a constant $\lambda$ to have the same scale as accuracy, but this will not change relative accuracy ranking since $PretrainedAcc$ in \eqref{perturb_a_q} is a constant.
As in \cite{dong2019hawqv2}, the total perturbation $Pert$ can be estimated by summing the perturbation contributed from each layer $Pert_i$. Using the norm of $\Delta W_i$ (the distance between the quantized tensor and the original tensor $W_i$) and the trace of Hessian matrix $H_i$, the $Pert_i$ can be calculated as follows (i is the layer index).

\begin{equation}
    \small
        \begin{gathered}
        Acc = PretrainedAcc - \lambda Pert, \\
        Pert = \mathbb P(A, Q) = \sum_{i=1}^{N} {Pert_i}, \\
        Pert_i = \overline{Tr}(H_i) \cdot \|\Delta W_i\|_2^2,
        \end{gathered}
    \label{perturb_a_q}
\end{equation}

With a latency constraint $Lat_0$, we need to find feasible neural architecture $A$ and then determine corresponding quantization setting $Q$ to minimize perturbation. Note that $A$ contains integer architectural parameters (kernel size, feature resolution, channel number, stride, etc), and $Q$ contains the bitwidths of layers in the subgraph, which are integer values chosen from \{2, 3, 4, 5, 6, 7, 8\}. Therefore, the task to find $A$ and $Q$ satisfying latency constraint $Lat_0$ can be formulated as an integer programming problem as shown in \eqref{integer_programming_formulation}.

\vspace{-2mm}
\begin{equation}
    \small
        \begin{gathered}
        \underset{Q}{\text{min}} ~ \mathbb P(A, Q), \\
        s.t. ~~ \mathbb L(A, Q) \leq Lat_0
        \end{gathered}
    \label{integer_programming_formulation}
\end{equation}

The latency constraint in \eqref{integer_programming_formulation} can be modified to \eqref{modified_latency_constraint} to reduce the number of neural architecture candidates. This modification is based on the assumption that neural architectures with higher latency tend to have more complex structures and higher expression capability, and therefore higher accuracy. $\alpha$ here is a hyperparameter ranging  from 0 to 1. A larger $\alpha$ can lead to a lower search cost.

\begin{equation}
    \small
        \begin{gathered}
        \alpha Lat_0 \leq \mathbb L(A, Q) \leq Lat_0
        \end{gathered}
    \label{modified_latency_constraint}
\end{equation}

We apply Monte Carlo tree search (MCTS)~\cite{kocsis2006bandit} for better sample efficiency on finding feasible neural architectures and quantization bitwidths that satisfy~\eqref{integer_programming_formulation} and~\eqref{modified_latency_constraint}. Benefiting from its online model, MCTS can dynamically trade off exploration and exploitation, which makes MCTS hard to be trapped in local optimum compared to other methods such as Bayesian optimization or greedy algorithms.
With the heuristic that $\mathbb L(A, 2bit) \leq \mathbb L(A, Q) \leq \mathbb L(A, 8bit)$, we first find $A$ that satisfies~\eqref{modified_latency_constraint_heuristic} and then solve for appropriate quantization setting $Q$. We follow the standard to set $A$ (then $Q$ in the next step) as state, and our actions are selected from \{increase/decrease channel, increase/decrease resolution, skip/unskip a layer, add/delete a subgraph, termination\}. More details about MCTS can be found in~\cite{kocsis2006bandit, auer2002finite, wang2019sample}.

\begin{equation}
    \small
        \begin{gathered}
        \alpha Lat_0 \leq \mathbb L(A, 8bit) \\
        \mathbb L(A, 2bit) \leq Lat_0
        \end{gathered}
    \label{modified_latency_constraint_heuristic}
\end{equation}

\subsubsection{Accuracy Predictor}
\label{subsec:accuracy_predictor}
As discussed in \sref{subsec:integer_programming}, given a latency constraint $Lat_0$, neural architecture candidates and corresponding quantization settings can be obtained with different perturbation. To compare among different neural architectures, a predictor is used to estimate the accuracy of pre-trained models with given architectures. In \OURS, we directly stack architectural parameters of each layer together as the input vector, and then we apply a support vector regression (SVR) model to predict the accuracy. It should be noted that we choose SVR predictor for simplicity and better sample efficiency, since SVR models generally require fewer data to train compared to neural networks used in \cite{wen2020neural, tang2020semi}.
To quickly train the predictor, we collect \{architecture, accuracy\} data by training 10 large neural networks from scratch and then reusing the weights while fine-tuning them to 200 different architectures. In our experiments, all neural networks are built by linearly stacking subgraphs, meaning that they are generally similar to each other. To support more complicated architectures such as DenseNet~\cite{huang2017densely} or LSTMs~\cite{sundermeyer2012lstm}, as suggested in \cite{wen2020neural, tang2020semi}, using a better strategy (such as autoencoder) for neural architecture representation, using semi-supervised learning with unlabelled data, and using graph convolutional networks (GCN) as the predictor can further improve performance, with the cost of more computation resources and time. 

We use the accuracy predictor to sort candidates that satisfy the latency constraint $Lat_0$. Since the accuracy predictor can be shared with different subgraphs, we repeat the aforementioned process for all subgraphs and select the top neural architectures and corresponding quantization settings\footnote{In our experiments we train top 5 architectures with corresponding quantization settings and choose the best one for a given latency constraint.}. We finally train them from scratch on ImageNet and then quantize the models as the final results of \OURS.

\section{Results}
\label{sec:results}
\subsection{Simulator Performance}

\begin{figure}[h]
\centering
\vspace{-9mm}
\includegraphics[trim=26 0 30 0, clip, width=.45\textwidth]{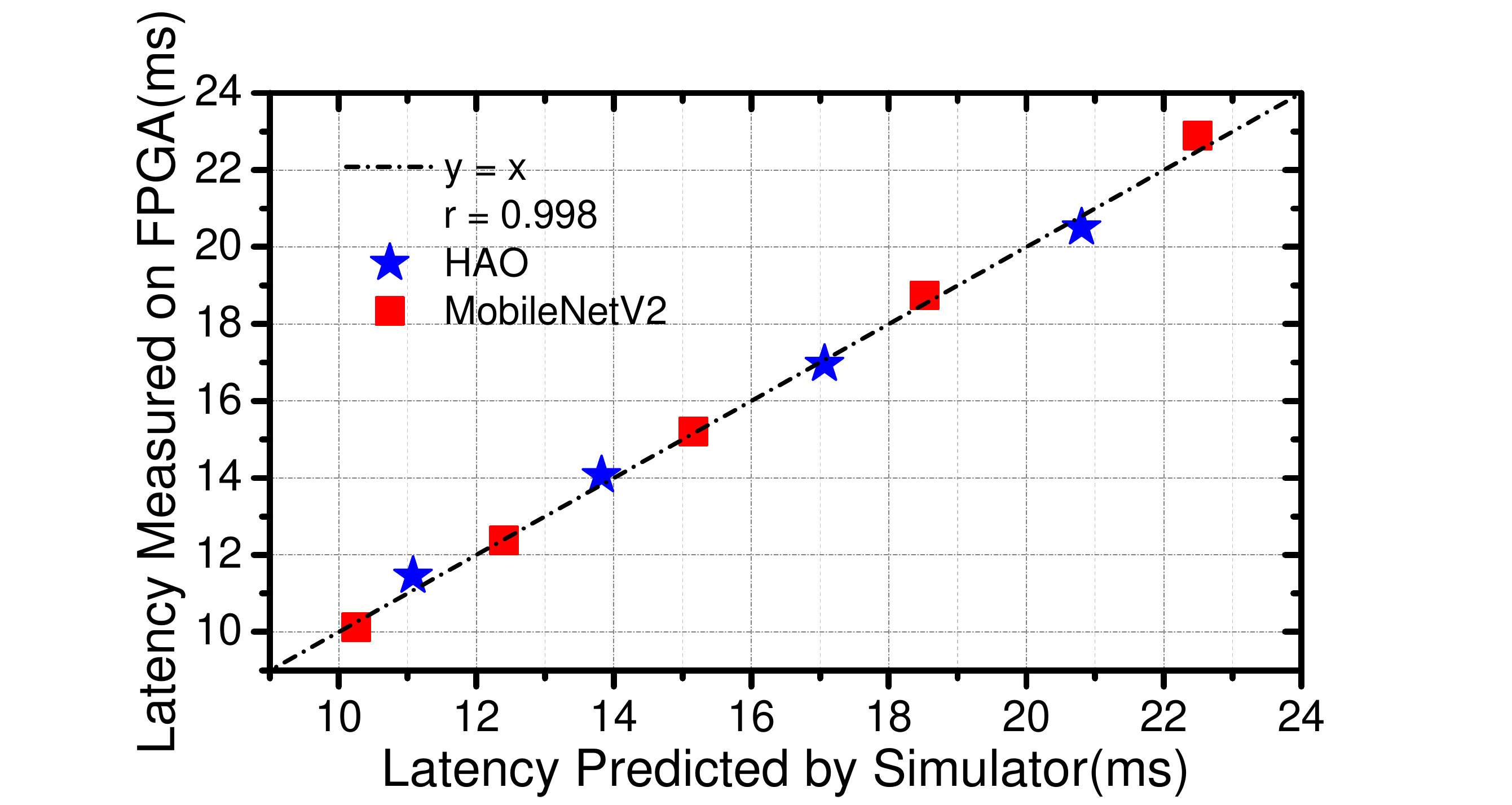}
\includegraphics[trim=26 0 30 20, clip, width=.45\textwidth]{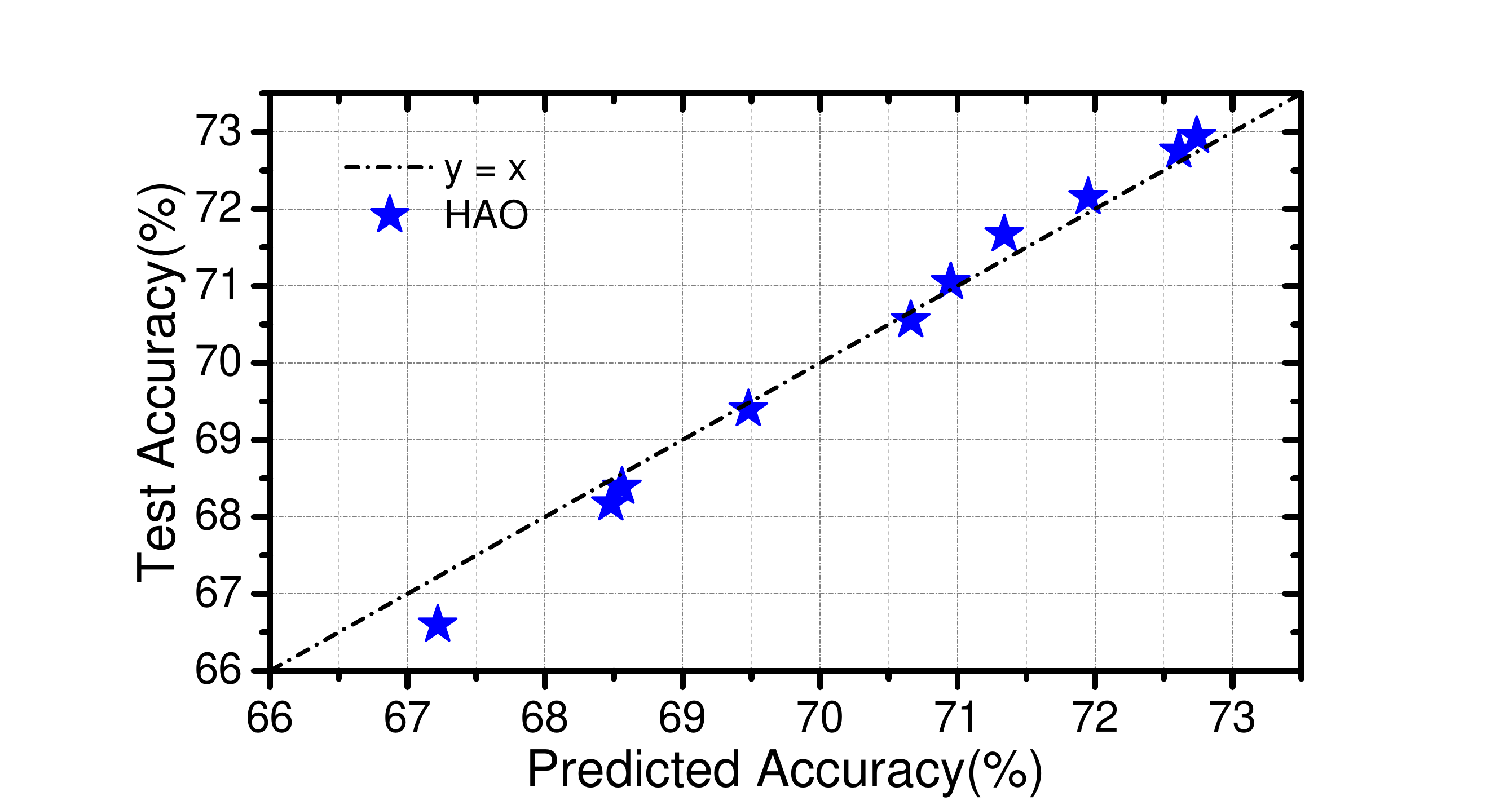}
\vspace{-2mm}
\caption{
\small
(Top) The correlation between latency predicted by the hardware simulator (after calibration) and the latency directly measured on FPGA. (Bottom) The correlation between predicted accuracy and the accuracy tested on ImageNet validation set.}
  \label{fig:correlation}
\end{figure}

In~\sref{subsec:hardware_design}, we present an analytical latency simulator that can quickly estimate the inference latency given a DNN architecture. The optimization algorithm in~\sref{subsec:integer_programming} uses the simulator to obtain quick latency feedback. 

To test the effectiveness of our latency simulator, we synthesize several accelerators for different MobileNetV2 and \OURS designs. The hardware parameters of different implementations are automatically generated by hardware optimization in~\eqref{eq:hardware_program}.
To calibrate our latency model for the target FPGA, we first perform linear regression to fit the cycle prediction to the hardware execution latency. We obtain a calibrated latency model $1.27 \times Lat+3.8$ and use it for our latency prediction.
Then for different accelerator implementations, 
we obtain the latency pairs from our simulator and the real hardware execution and plot them in ~\fref{fig:correlation}.
 We observe a strong linear relationship ($r=0.998$) between the real inference latency and the estimated latency.

In addition to the hardware latency simulator, \OURS also uses an accuracy predictor to reduce the computational cost.
We show the performance of the predictor in~\fref{fig:correlation}. 
As can be seen, for different CNN models in our search space, the results of our accuracy predictor align well with the actual test accuracies on ImageNet validation dataset.

\subsection{Experimental Results}
In this section, we present the accuracy and latency results of \OURS on the Ultra 96 board with a Xilinx Zynq ZU3EG FPGA. We show that \OURS outperforms manually designed solutions, as well as solutions with automatically searched DNN architectures and quantization settings.

\begin{figure}[t]
    \centering
    \vspace{-3mm}
    \includegraphics[width=.43\textwidth]{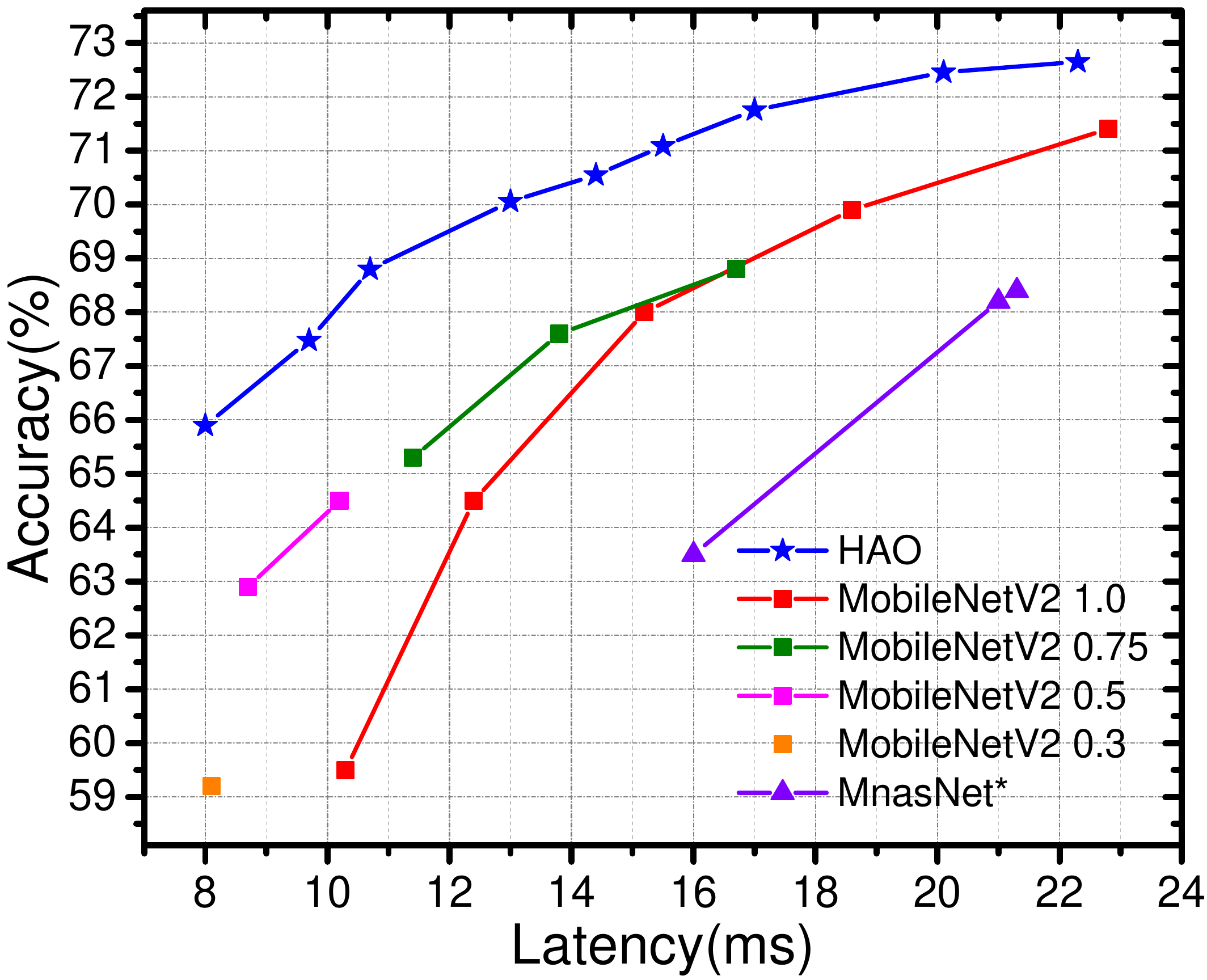}
    \vspace{-3mm}
    \caption{
    \small
    Pareto frontier for accuracy and latency. We generate pareto frontier of MobileNetV2 and MnasNet by varying width multipliers as well as the input resolution, as suggested in the references~\cite{sandler2018mobilenetv2, tan2019mnasnet}. As can be seen, \OURS results outperform MobileNetV2 and MnasNet by a large margin on Zynq ZU3EG.
    }
      \label{fig:pareto_curve}
    \end{figure}

\vspace{1mm}
\fref{fig:pareto_curve} shows the pareto frontier of \OURS with respect to accuracy and latency. MobileNetV2 \cite{sandler2018mobilenetv2} is a popular neural architecture manually designed for efficient inference. The original MobileNetV2 is in floating-point format. To achieve a fair comparison, we quantize MobileNetV2 to 8-bit weights and 8-bit activations, and then run it on FPGA with a \{1x1 convolution, 3x3 depthwise convolution, 1x1 convolution\} subgraph. We follow \cite{sandler2018mobilenetv2} to change the channel width multiplier (selected from $\{1.0, 0.75, 0.5, 0.3\}$) and input resolution (selected from $\{224, 192, 160, 128, 96\}$) of MobileNetV2, in order to trade-off latency and accuracy. In comparison, the neural architecture (including input resolution) and quantization bitwidth setting are automatically selected in \OURS. As can be seen, \OURS outperforms MobileNetV2 on a wide range of latency values. \OURS can achieve 72.5\% top-1 accuracy with 20ms latency (50 fps), which is more than 1\% higher accuracy than MobileNetV2 while running 15\% faster. In the cases with a more strict latency constraint (for example autonomous vehicles), \OURS can still preserve 66\% accuracy with only 8ms latency (125 fps). This is significantly higher than the 63\% of MobileNetV2 while being faster. Furthermore, we compare with results from MnasNet~\cite{tan2019mnasnet}, which is a hardware-aware neural architecture search method. As in~\fref{fig:pareto_curve}, \OURS also outperforms MnasNet by a large margin\footnote{Part of the MnasNet pareto curve is out of the latency range in~\fref{fig:pareto_curve}. We present these extra results in~\tref{tab:time}.}.

\begin{table*}[h]
    \centering
\caption{Performance comparison on ImageNet with prior works.}    
\vspace{-2mm}
\begin{tabular}{|l|l|l|l|l|l|r|}
\hline
\textbf{}   & \textbf{Platform} & \textbf{Input Resolution} & \textbf{Framerate(fps)} & \textbf{Quantization Bitwidth} & \textbf{Top-1 Accuracy(\%)} \\ \hline
EDD-Net-2~\cite{li2020edd} & Zynq ZU9EG & 224 $\times$ 224 & 125.6               & W16A16 & 74.6  \\
HotNas-Mnasnet~\cite{jiang2020standing} & Zynq ZU9EG & 224 $\times$ 224 & 200.4  & NA      & 73.24 \\
HotNas-ProxylessNAS~\cite{jiang2020standing} & Zynq ZU9EG & 224 $\times$ 224 & 205.7    & NA & 73.39 \\ 
EDD-Net-3~\cite{li2020edd} & Zynq XC7Z045 & 224 $\times$ 224 & 40.2              & W16A16 & 74.4  \\
VGG16~\cite{zhang2018dnnbuilder} & Zynq XC7Z045 & 224 $\times$ 224 & 27.7        & W16A16 & 69.3  \\
VGG-SVD~\cite{qiu2016going} & Zynq XC7Z045 & 224 $\times$ 224 & 4.5              & W16A16 & 64.64 \\
VGG16~\cite{suda2016throughput} & Stratix-V & 224 $\times$ 224 & 3.8             & W8A16  & 66.58 \\
VGG16~\cite{guo2017software} & Zynq 7Z020 & 224 $\times$ 224 & 5.7               & W8A8   & 67.72 \\
Dorefa~\cite{jiao2017accelerating} & Zynq 7Z020 & 224 $\times$ 224 & 106.0       & W2A2   & 46.10 \\
Synetgy~\cite{yang2019synetgy} & Zynq ZU3EG & 224 $\times$ 224 & 66.3            & W4A4   & 68.30 \\
FINN-R~\cite{blott2018finn} & Zynq ZU3EG & 224 $\times$ 224 & 200.0              & W1A2   & 50.30 \\
MobileNetV2~\cite{sandler2018mobilenetv2} & Zynq ZU3EG & 224 $\times$ 224 & 43.5 & W8A8   & 71.40 \\
MnasNet-A1~\cite{tan2019mnasnet} & Zynq ZU3EG & 224 $\times$ 224 & 22.3          & W8A8   & 74.60 \\
MnasNet-A1~\cite{tan2019mnasnet} & Zynq ZU3EG & 192 $\times$ 192 & 27.8          & W8A8   & 73.33 \\
MnasNet-A1-0.75~\cite{tan2019mnasnet} & Zynq ZU3EG & 224 $\times$ 224 & 31.0     & W8A8   & 72.70 \\
MnasNet-A1~\cite{tan2019mnasnet} & Zynq ZU3EG & 160 $\times$ 160 & 35.8          & W8A8   & 71.35 \\
FBNet-B~\cite{wu2019fbnet} & Zynq ZU3EG & 224 $\times$ 224 & 24.6                & W8A8   & 73.20 \\
FBNet-iPhoneX~\cite{wu2019fbnet} & Zynq ZU3EG & 224 $\times$ 224 & 21.3          & W8A8   & 72.62 \\

\rowcolor{Gray}
\textbf{\OURS}& Zynq ZU3EG  & 256 $\times$ 256 & 44.9 & W-mixed A8 & 72.68 \\
\rowcolor{Gray}
\textbf{\OURS}& Zynq ZU3EG  & 256 $\times$ 256 & 50.0 & W-mixed A8 & 72.45 \\
\rowcolor{Gray}
\textbf{\OURS}& Zynq ZU3EG & 224 $\times$ 224 & 58.9  & W6A8    & 71.76 \\ 
\rowcolor{Gray}
\textbf{\OURS}& Zynq ZU3EG  & 224 $\times$ 224 & 77.0 & W-mixed A8 & 70.06 \\ 
\rowcolor{Gray}
\textbf{\OURS}& Zynq ZU3EG  & 192 $\times$ 192 & 93.5 & W-mixed A8 & 68.80 \\
\hline
\end{tabular}
\label{tab:time}
\vspace{-3mm}
\end{table*} 

    \begin{figure*}[]
    \centering
    \includegraphics[trim=0 278 0 100, clip, width=0.95\textwidth]{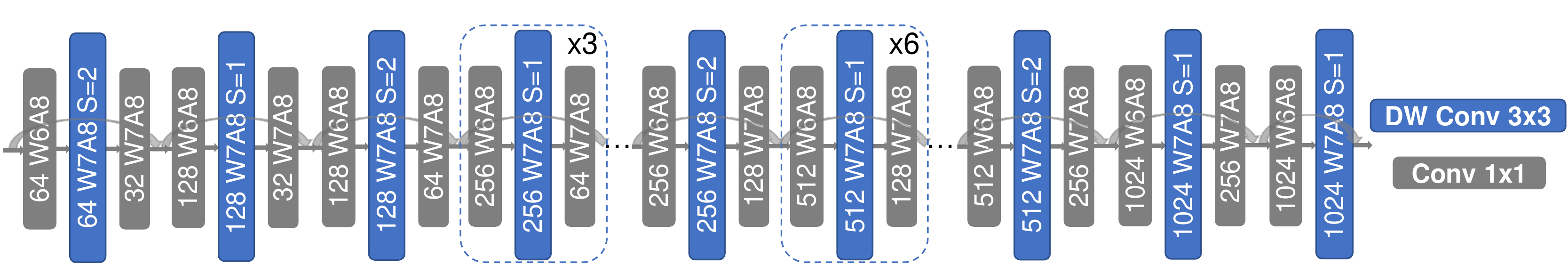}
    \vspace{-2mm}

    \caption{
    \small
    Illustration of neural architecture and quantization setting searched by \OURS. W and A stand for weight and activation quantization bitwidth, S is the stride of a specific convolutional layer. DW Conv stands for depth-wise convolution.}
      \label{fig:model_illustration}
    \end{figure*}

In addition to comparing pareto-frontier performance with our own hardware implementation, we also compare \OURS with various previous work in \tref{tab:time}. \cite{qiu2016going, jiao2017accelerating, guo2017software, suda2016throughput} are manually designed solutions. \cite{li2020edd, jiang2020standing} are search-based methods. Note that these prior works target larger FPGA boards with more resources, and some use more complex neural architectures, 16-bit fixed-point or floating-point precision. For a fair comparison, we further compare \OURS with \cite{yang2019synetgy, blott2018finn, tan2019mnasnet, wu2019fbnet, sandler2018mobilenetv2}, which have the same hardware platform (Zynq ZU3EG) as ours\footnote{Note that \cite{tan2019mnasnet, wu2019fbnet} are well-known hardware-aware search algorithms, and we implement their searched results on Zynq ZU3EG for comparison.}. For \OURS, we apply layer-wise quantization for activations and channel-wise quantization for weights, with standard linear quantizer and static quantization 
for the simplicity of deployment. As can be seen in~\tref{tab:time}, \OURS achieves state-of-the-art performance on embedded FPGA with limited resources. With higher top-1 accuracy (68.8\% vs 68.3\%), \OURS solution is significantly faster than Synetgy~\cite{yang2019synetgy} (94fps vs 66fps), albeit Synetgy is assisted by extra operations such as shift. Moreover, when the framerate is 50fps, \OURS can achieve 72.5\% top-1 accuracy on ImageNet, which is more than 1\% higher than MnasNet-A1 (71.4\%) while being 14\% faster. Comparing with FBNet-iPhoneX, \OURS obtains slightly better accuracy (72.7\% vs 72.6\%), while having a much higher framerate (45 vs 21). 
It should be noted that for different hardware platforms or different latency constraints, previous methods need to repeat the whole search pipeline to find appropriate solutions, while the predictor in \OURS can be shared so that no additional search cost will be required.

\begin{table}[htp]
\centering
\caption{Hardware resources utilization and power}
\vspace{-1mm}
\begin{tabular}{|c|c|c|c|c|}
\hline
    LUTs          & FF            & DSP        & BRAM       & Power\\
\hline
    61362(87.0\%) & 55136(39.0\%) & 360(100\%) & 431(99.8\%)& 5.5W\\
\hline
\end{tabular}
\label{tab:hardware_utilization}
\end{table}

~\tref{tab:hardware_utilization} shows the hardware resource utilization and power usage for \OURS on Zynq ZU3EG FPGA. We observe 4.3W power consumption with no workload running on the programming logic side and 5.5W power when running the network. Besides, we are able to utilize 100\% of DSP and 87\% of LUTs on the FPGA, showing the effectiveness of our hardware resource modeling. In the optimization program in \eqref{eq:hardware_program}, we allocate $\beta$ percent of LUTs as computation resource to search for optimal design parameters, which makes the LUTs utilization more controllable. In this way, the simulator can automatically decide whether to implement a kernel on DSP or LUTs based on the quantization setting $Q$. As a result, we can achieve high resource utilization by leveraging the benefits of mix-precision operations on FPGA.

In~\fref{fig:model_illustration}, we show one of the searched results by \OURS.
A subgraph \{1x1 convolution, 3x3 depthwise convolution, 1x1 convolution\} is used in this solution. As can be seen, \OURS finds that a 6-bit/7-bit mixed-precision quantization setting is better than 8-bit uniform quantization for weights. 
In general, lower bit-width means more computation units under the same resource constraints, but it can lead to larger quantization perturbation. \OURS can balance the efficiency and perturbation, and we observe that the 8-bit counterpart of \OURS 6/7-bit result runs 5\% slower with negligible accuracy gain.
Moreover, the results of \OURS show that, for our implementation on Zynq ZU3EG, solutions with solely $3\times3$ depthwise convolution perform better than those with a mixture of $3\times3$ and $5\times5$ depthwise convolution. This is due to the fact that when using a mixture of $3\times3$ and $5\times5$ depthwise convolution, either $3\times3$ or $5\times5$ kernel will be idle when invoking the accelerator, which is a waste on platforms with limited hardware resources.

\section{Conclusions}
\label{sec:conclusions}
In this work, we propose \OURS to jointly optimize the neural architecture, quantization, and hardware design. %
To reduce the computation required for evaluating different designs,  
we develop a subgraph-based hardware latency model as well as an accuracy predictor for neural architectures. We formulate the algorithm and hardware co-search as an integer programming problem, which significantly prunes the total search space. 
On an embedded FPGA device, we show that our \OURS method finds the pareto-optimal designs which outperform previous solutions on both latency and accuracy.

\section*{Acknowledgments}
\small{
This work was supported by Facebook Reality Labs, Google Cloud, Alibaba, Samsung SAIT, 
by the Berkeley ADEPT Lab, Berkeley Deep Drive, the Berkeley Wireless Research Center, by the Croucher Innovation Award, 
and by CONIX Research Center. 
}

\clearpage
{
\bibliographystyle{ieee_fullname}
\bibliography{egbib}
}

\end{document}